# TOWARDS AN INTELLIGENT SYSTEM FOR RISK PREVENTION AND EMERGENCY MANAGEMENT

**Fahem Kebair**       **Frédéric Serin**

Laboratoire d'Informatique de Traitement de l'Information et des Systèmes
University of Le Havre
{fahem.kebair, frederic.serin}@univ-lehavre.fr

**ABSTRACT**

Making a decision in a changeable and dynamic environment is an arduous task owing to the lack of information, their uncertainties and the unawareness of planners about the future evolution of incidents. The use of a decision support system is an efficient solution for this issue. Such a system can help emergency planners and responders to detect possible emergencies, as well as to suggest and evaluate possible courses of action to deal with the emergency. We are interested in our work to the modelling of a monitoring preventive and emergency management system, wherein we stress the generic aspect. In this paper we propose an agent-based architecture of this system and we describe a first step of our approach which is the modeling of information and their representation using a multiagent system.

**INTRODUCTION**

The subject of crisis and emergency management continues to be a large scientific challenge due to many factors: natural, industrial, human, etc. The circumstances change from a case to another and from a situation to another. The necessity to adapt to these circumstances and to the specific issues that involve is the key of the emergency management efficiency. An emergency situation itself is dynamic and changes continuously, thereby it is arduous for planners and responders to produce robust plans towards both short-term and long-term goals. Information systems for risk prevention and emergency management are very useful verily indispensable to support decision-makers to manage emergency situations. One approach to address this challenge is to develop Decision Support Systems (DSS) which incorporate cognitive-level models of decision making and which can assist decision makers in real-time compilation of new procedures.

We aim through this research to bring our contribution in the crisis and emergency management field. Our objective is to develop a monitoring preventive and emergency management system we want the most generic possible. Our motivation to build a system as generic as possible is based on the observation that the current several applications proposed so far are dedicated to particular cases and are not or hardly adaptive to new subjects of studies. We think the multiagent approach is then the most suitable technology to endow the system with adaptivity and flexibility abilities. In this paper we propose an agent-based architecture of a DSS we consider generic for the most part. The work presented here is dedicated to the RoboCupRescue Simulation System (RCRSS) (Kitano, Tadokor, Noda, Matsubara, Takhasi, Shinjou, and Shimada, 1999) in which we focus on a first step of our approach which is the modeling of information about the observed environment and their dynamic representation using a multiagent system.

The paper is structured as follows: first, a survey of information technologies and systems for the emergency management is discussed. Then, an overview of our research objective and the approach we are going to deal is presented. Next, the internal architecture of the emergency management system is described. After that, a description of the used approach for the modelling and the representation of information, as well as an implementation and graphic tools are provided. Finally, a conclusion and perspectives are mentioned.

**SURVEY OF INFORMATION TECHNOLOGIES AND SYSTEMS FOR THE EMERGENCY MANAGEMENT**

Information Technologies (IT) and advanced modeling techniques continue to help society to limit and manage crisis incidents (Rinaldi, Peerenboom, and Kelly 2001). Indeed, IT can help in disaster response since improving the information management during the disaster–collecting information, analyzing it, sharing it, and disseminating it to the right people at the right moment–will improve the response by helping humans make more informed decisions. A number of research efforts have been targeted at the topic of emergency and disasters management with the aim to create modeling and simulation techniques and tools for the emergency management. A classification of these tools is described in (Jain, 2006). Such tools may be used on the occurrence of an actual event, or in preparedness role, that is, for planning the response for future potential incidents. Risk detection and prevention was also discussed in the works of Beroggi and Wallace (Beroggi and Wallace 1994; Beroggi et al. 2000) in which dynamic and adaptive models for operational risk management are proposed. These models allow monitoring process, evaluating, and changing courses of action with potential detrimental consequences in real time.

DSSs are one of the technologies used for the emergency management and that demonstrates their abilities to support and improve decision-making. However, flexibility remains crucial to the success of planning and response operations (Mileti, 1999; Stewart and Bostrom, 2002), and to be successful such a system needs to be adaptive, easy to use, robust and complete on important issues (Little, 1970). During the last decade, intelligent agents and Multiagent Systems (MAS) technologies are considered to be a promising method to construct the scalable, robust, reusable high quality software system thanks to agents characteristics that can be summarized as follows: proactivity, reactivity, social ability, intelligence and purpose (Graesser and Franklin, 1996). Articles on the future directions of DSS research have stressed the importance of using agent technologies in DSS (Carlsson and Turban, 2002; Shim, Warkentin, Courtney, Power, Sharda, and Carlsson, 2002). Agents embedded in such systems can decide on information being monitored and filtered in geographically distributed environments (Bui and Lee, 1999), and can gather information to promote user awareness (Budzik, Bradshaw, Fu, and Hammond, 2002) and support decision-making (e.g., use airline databases to support travel planning (Nunes-Suarez, Sullivan, Brouchoud, Cross, Moore and Byrne, 2000).

Another aspect to which we are interested in our work is the genericity. To test and to prove this genericity, we are working currently on several applications in relatively varied domains. A first prototype dedicated to the game of Risk was built (Person, Boukachour, Coletta, Galinho and Serin, 2006). Then, we started working recently on the RoboCupRescue (RCR) project for which we are developing currently a new prototype version (Kebair, Serin and Bertelle, 2007). Finally, we began a collaboration on E-learning with specialists in didactics (Bertin and Gravé, 2004).

**PROPOSED APPROACH**

**Research objective and approach overview**

The approach discussed here is addressed to the modeling and the design of a monitoring preventive and emergency management system we consider generic for the most part. The problem we are dealing with concerns partially perceived environments for which a preliminary and static knowledge is acquired, which is completed by dynamic and permanently observations. These observations may be partially and incomplete: each observed entity is not exhaustively described and all the entities participating to this observation are neither described.

The objective is to detect upstream a risk of dysfunction namely to perceive in time that a situation is liable to change into a crisis verily into a catastrophe. We have to address the detection of significant organisations that provide a meaning to data to help finally to decision-making.

Another aspect of our methodological approach is to formalise not the environment itself that may be extremely different according to the treated subjects and that, by definition derived from the viewpoint of the systemic analysis can not be clearly explicit, but to model the observations issued from this environment. These observations are therefore reified according to an object model. This formalisation represents a framework that help to build an ontology whereof construction is often strenuous and requires an expertise of the studied domain. Furthermore, this reification is used to specify the generic structure of the Factual Semantic Features (FSF) in which will be written the observations reaching the system. Then, thanks to the ontology of the domain we are able to define the content of these FSFs. The latter will be presented later in this paper.

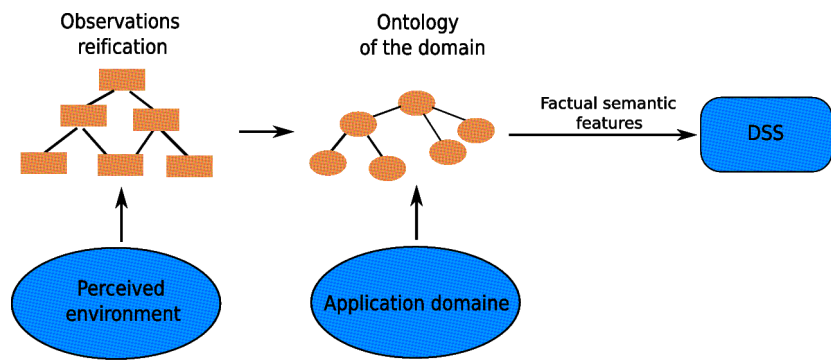

**Figure 1. First step in the decision-making process**

**System internal mechanism**

Figure 2 shows the whole architecture of the emergency management system. The system is feed permanently by information sent by actors and which describe the current state of the environment. Information are handled thereafter by agents which form a part of the system kernel. The system needs some knowledge about the environment as the ontologies of the domain and the proximity measures. Outcomes provided by the system include an evaluation of the situation and an emergency management plan to manage it.

The system has a layered multiagent kernel. A three level process is integrated in our model involving the representation, the characterisation and the interpretation of factual observations. As follows the internal mechanism of the system:

- The first step is to deal with information coming from the environment thanks to factual agents. These agents form a *representation layer* that has as essential role to represent dynamically the current situation. Each factual agent aims to reflect a partial part of the observed situation. Its objective is to reach a predominant place in the factual MAS by fighting some agents and helping some others.

- The second step is to analyse the emergent organisation of factual agents by creating agents in a *characterisation layer*. This MAS is generated using dynamic clustering techniques. Each cluster of factual agents leads to the creation of a characterisation agent (or clustering agent). This agent represents by fact which emerges by similarity between diverse factual agents and dissimilarity between characterisation agents. This clustering is not necessarily supervised since one of our objectives is to detect a risk not inevitably referenced as such; it is dynamic because the observed entering facts modify permanently the structure of the factual MAS and dysfunctions may evidently appear during the observation or, on the contrary disappear.

- The third step constitutes the fundamental phase of the system decision-making process where scenarios are identified and carried by prediction agents. These agents constitute a *prediction layer* and aim to find clusters, in the characterisation layer, enough close to inform the decision-makers about the current situation and its probable evolution, verily to generate a warning in the case of detecting a risk of crisis. This mechanism is managed by a Case Based Reasoning (CBR) and is studied "manually" by the expert of the case study. The CBR must provide the possible consequences of a given scenario and the solution that match to this particular case of the situation.

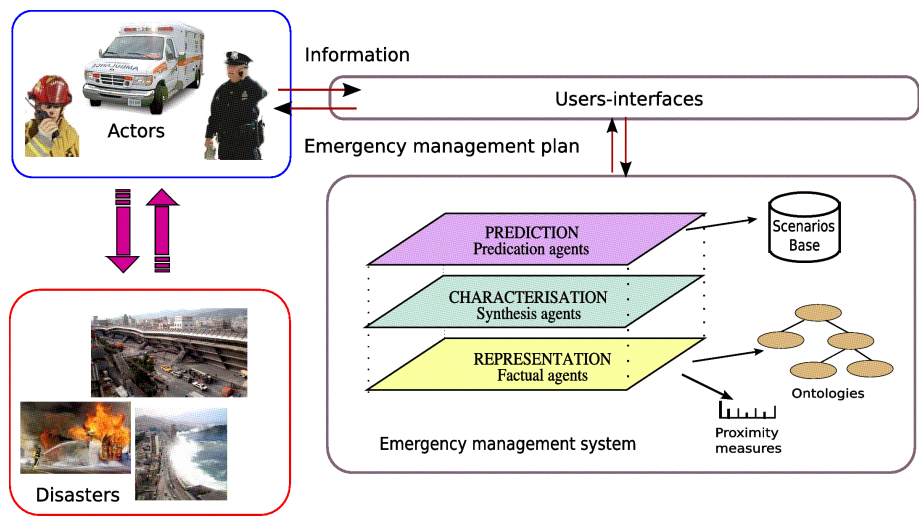

**Figure 2. Proposed architecture of the emergency management system**

## DESIGN OF THE REPRESENTATION LAYER

### Case-study: RoboCupRescue Simulation System

The RCRSS is an agent-based simulator which intends to reenact the rescue mission problem in real world. An earthquake scenario is reproduced in the RCR environment including various kinds of incidents as the traffic after earthquake, buried civilians, road blockage, fire accidents.

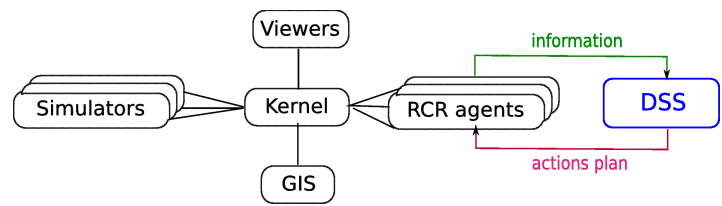

**Figure 3. Interaction of the DSS and the RCR agents**

A set of heterogeneous agents coexist in the disaster space, each with a specific goal and a particular role. We distinguish civilian agents that must be saved, platoon agents that are the rescue teams (or RCR agents) as fire brigade, police force and ambulance teams and center agents that are the three command centers of the platoon agents. Our contribution in this project is to design the rescue teams and to improve their decision making abilities using the DSS. Figure 3 shows the architecture of the RCRSS and the interaction of RCR agents with the DSS. The latter receives perceived information sent by RCR agents and sends conversely an action plan previously set up according to a set of preset scenarios.

**Observations reification**

A reification of the observed environment using the object paradigm is needful to format the observations in FSFs. Each observation is dual and includes a description, at the same time, of physical entities and a dynamic of the environment. From diverse concrete cases we propose a decomposition in six classes that may be qualified as abstract or generic (Figure 4). These classes belong to two families or-in other words-inherit two super classes. The first one is constituted by individual physical objects or exceptionable composite; it gathers the three classes that are passive entities or objects of the environment, actors which are active entities and means that are collection of various physical objects. These classes describe therefore direct and concrete observations concerning passive and active entities; means are introduced to characterise a set of observable objects of which we could specify related behaviours such as the changes of correlated states, concentrated behaviours or physical grouping that infer to emergent and specific actions of this gathering.

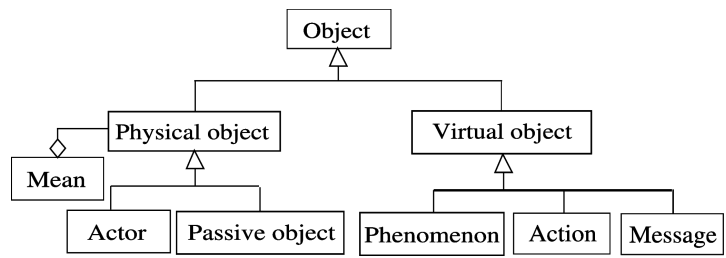

**Figure 4. Observations object model**

The second family categorises the virtual objects which are the different forms of activities; it gathers the three other classes which are phenomena, actions and messages. The elements of this family are deduced from indirect observations and are formalised basing-on the "memento" design pattern of Gamma (Gamma, Helm, Johnson, and Vlissides, 1995). They are differentiated by their interconnections with physical objects. Actions are related to actors or means, which are filled with goals, and take action on the states of other physical objects.

Phenomena have no goals for the least intelligible neither actors that trigger them. Messages are very close to actions and are distinguished by the lack of change in states without intermediate pretreatment verily without any notable change.

**Factual Semantic Features and Ontology**

Formatting the reified observations according to the object-modelling and to the ontology of the domain introduces for our system the fundamental notion of FSFs. The given noun to this message content brings an account to our approach: we stress observed and punctual elements that are facts namely observed entities directly and beforehand deduced and that we estimate to be bearer of meaning for the grasp, the appropriation and the analysis of the environment. FSFs are initial elements that permit to detect risks. They are the messages content referring to the FIPA denomination (FIPA), they join other components that are the source of this feature and the various parameters such as performative and relevance that we do not describe here. Each FSF is composed of an object that describes, issued from the object-model presented above. To each object, the FSF associates qualifiers and their related values. The ontology presented in Figure 5, implemented using protégé (Protégé), defines the set of the observable entities and their corresponding qualifiers in the RCR environment. Two categories of objects will be carried by FSFs:

a. RCR agents which are the actors of the environment. These objects inherit PlatoonAgent class and include FireBrigade, PoliceForce and AmbulanceTeam objects. For example, an FSF related to a FireBrigade object is:

(fireBrigade#5, hit point, 100, fires, 2, team, 3, action, extinguish, target, building#5, localisation, 7|9, time, 5)

This FSF means a fire brigade is extinguishing building#5 at the $5^{th\ a}$ cycle of the simulation and is localised by the 7|9 coordinates. The hit point indicates its health state (the agent is dead when the hit point is equal to 0), has two fires around it and is supported by three team-mates.

b. Phenomena objects ineherit Phenomenon class and include the following objects: Fire, Collapse, Injury and Blockade. An example of an FSF related to a Fire object is:

(fire#14, fieriness, 1, inDangerNeighbours, 3, burningNeighbours, 2, localisation, 20|25, time, 7)

This FSF means a fire started in building#14, its degree is equal to 1 and was perceived in the $7^{th}$ cycle of the simulation. The fire has 2 neighbour fires, can spread on 3 unburned neighbours buildings and has the following coordinates 20|25.

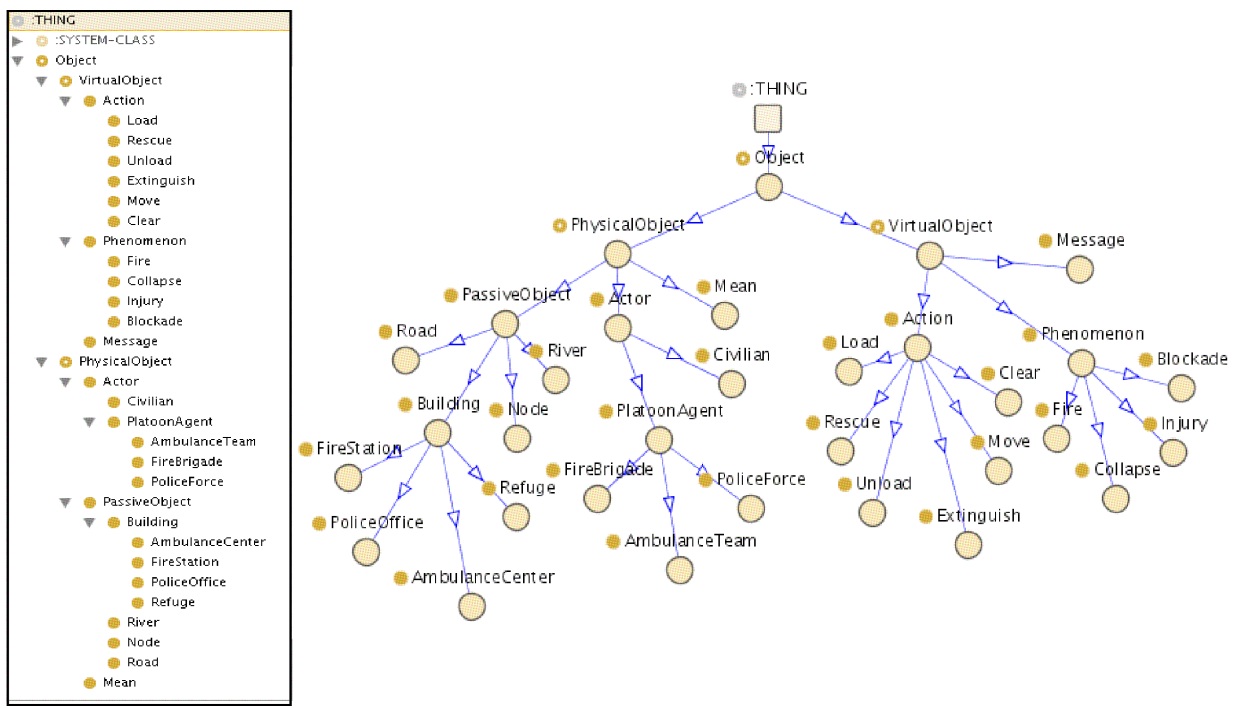

**Figure 5. Ontology of RoboCupRescue simulation environment**

In order to make emerge the significant facts of the current situation, we have introduced the proximity notion between FSFs. Three types of proximities are distinguished: temporal proximity (*Pt*), spatial proximity (*Pe*) and semantic proximity (*Ps*). We can speak about temporal and spatial distances. The more two events are distant in

---

[a] A cycle represents one second in the simulation

time and space, the smaller the proximity is. *Pt* and *Pe* are computed using a sigmoid function, it takes into account the negative values and is defined to remain in an interval of [0,1]. It brings the five following advantages: its continuity, its derivability, the knowledge of its primitive, its definition on $\mathcal{R}$ entire (including negative values) and its symmetry in zero. The following formula compute respectively *Pt* and *Pe*:

$$Pt = \frac{\left(4e^{-\Delta t}\right)}{\left(1+e^{-\Delta t}\right)^2}$$

where $\Delta t$ is the difference of time

$$Pe = \frac{\left(4e^{-\Delta e}\right)}{\left(1+e^{-\Delta e}\right)^2}$$

where $\Delta e$ is the euclidean distance between two objects

The definition of *Ps* is related to the definition of the ontology. The semantic proximity here functions like a similarity. The total proximity between two FSFs provides a value in [-1,1] and is computed by this formula: $Pe * Pt * Ps$. Two FSFs are similar if they have a proximity equal to 1; they are opposite when the proximity is equal to -1; and are neutral or incomparable if their proximity is equal to 0.

**Factual Agents**

The acquire of the FSFs sent to the system is insured by the representation layer which is composed of factual agents. The kind of this treatment is directly resulting from the proximity value between the diffused observation and the information already contained in each agent. In order to not loss any information, a *generative* agent creates a new factual agent that will incorporate the original information if the approbation by existent agents is estimated insufficient (effect of semantic proximities threshold).

Each factual agent has an internal automaton which is an Augmented Transition Network (ATN). The role of the ATN is to manage the behaviour and to describe the lifecycle of its agent. The ATN has a generic structure. However, the number of states and the set of conditions and actions carried by the transitions are specific to the factual agent to which the automaton is attached.

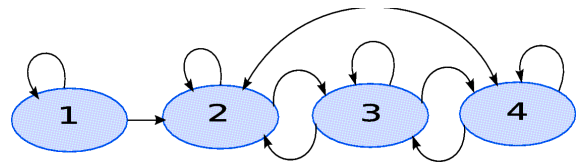

**Figure 6. Generic ATN**

A generic acquaintances network is used by each factual agent in order to interact with other agents. The agent stores in this network the list of the agents with which it has a proximity different from 0. Factual agents has also indicators that reflect their dynamic in the representation MAS and that provide thereby a synthetic view of the salient facts of the situation. They depend on the processed application, and their meanings may differ from a factual agent type to another. Two indicators are defined for the factual agents in RCR:

*Action indicator (AI)*: it represents the position and the strength of a factual agent inside the representation MAS. For factual agents related to RCR agents, *AI* means the potential of an RCR agent and its efficiency in solving a problem. For factual agents managing phenomena, *AI* means the degree of damage and hazard that could represent this phenomenon.

*Probability indicator (PI)*: For factual agents related to RCR agents, *PI* means the ability of an RCR agent to discover new problems in the disaster space. For phenomena factual agents, *PI* means the solving probability and the worsening impediment of a phenomenon.

**IMPLEMENTATION AND GRAPHIC TOOLS**

The implementation of the representation layer is underway. The tests carried out so far include a part of the ontology dealing with fires. We choose JADE (Bellifemine, Caire, Trucco, and Rimassa, 2007) platform to implement the representation MAS because it is FIPA compliant, but also because it is easy to implement factual agents using JADE agents.

Figure 7 shows an ATN of a factual agent related to a fire object. We distinguish four states: a *Creation* state in which the agent is created; a *Burning* state in which the fire is burning and can spread on other unburned neighbour buildings; a *Put out* state where the fire is in an extinguish phase; and an *Off* state that means fire ended.

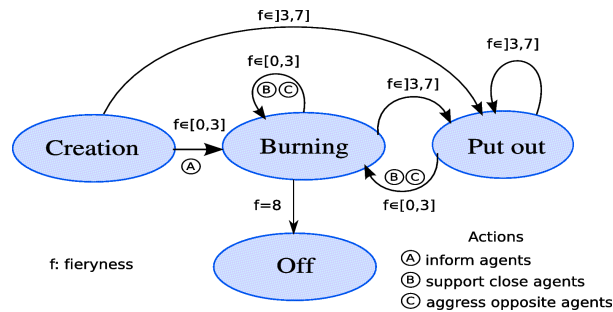

**Figure 7. ATN of a fire factual agent**

A factual agent related to a fire changes state according to its fieriness value. The agent performs actions when it changes state. For example, it informs all agents about its creation from *Creation* state to *Burning* state, supports close agents and aggress opposite agents from *Burning* state to *Put out* state. In *Off* state the agent is considered as dead and ceases activity, since the information that it represents has not any importance, and consequently the related factual agent does not have anymore function to fulfill in the representation MAS.

We have created and tested some specific graphic tools in order to analyse the representation MAS. We use an interactive interface (Figure 8) to read in real time information about the RCR disaster space and factual agents properties: their FSFs, their states, their indicators and their acquaintances networks. To have a static view of the representation MAS at a given moment of the situation, we freeze all the agents. Thus, we may have a finer analysis of the factual agents. Moreover, this allows us to evaluate the way we form clusters in the second layer, or more precisely the way we compare between the emergent clusters and the stored scenarios.

Figure 9 shows a partial part of the RCR disaster space (on the left) at the $31^{th}$ of the simulation. We use a second tool (on the right) to analyse the behaviour and the evolution of all the created factual agents. The first column is the name of fires factual agents. The next four columns are the possible states of the internal automaton. States 1, 2, 3 and 4 refer respectively to *Creation*, *Burning*, *Put out* and *Off* states of a fire factual agent. The last column is used to display the two values of the indicators AI and PI. The evolution of these values are displayed inside the cells respectively with red and blue colors. Here, we show some examples of factual agents that are linked to their corresponding fires. The two agents f#129769672 and f#268425221 represent two fires in an extinguish stage. Consequently, they are in state 3 with high values of PI and with values of AI close to 0. Agent f#268425221 is burning with a fieriness value equal to 1, that means it will be extinguished soon and it no longer represents a danger since all its neighbourhood is burning. It has therefore a negative value of PI (represented with purple color) and a low value of AI. The fire decreases from an importance point of view, what will influence the fire brigade decision-making that relies essentially on the choice of fires with highest priority, accordingly important. This is the case of agent f#266324026 which represents an important fire, since it has a fieryness value equal to 1, is at its beginning and may spread on its neighbours. It has therefore positive and relatively high values of AI and PI.

**Figure 8. Agents interactive interface**

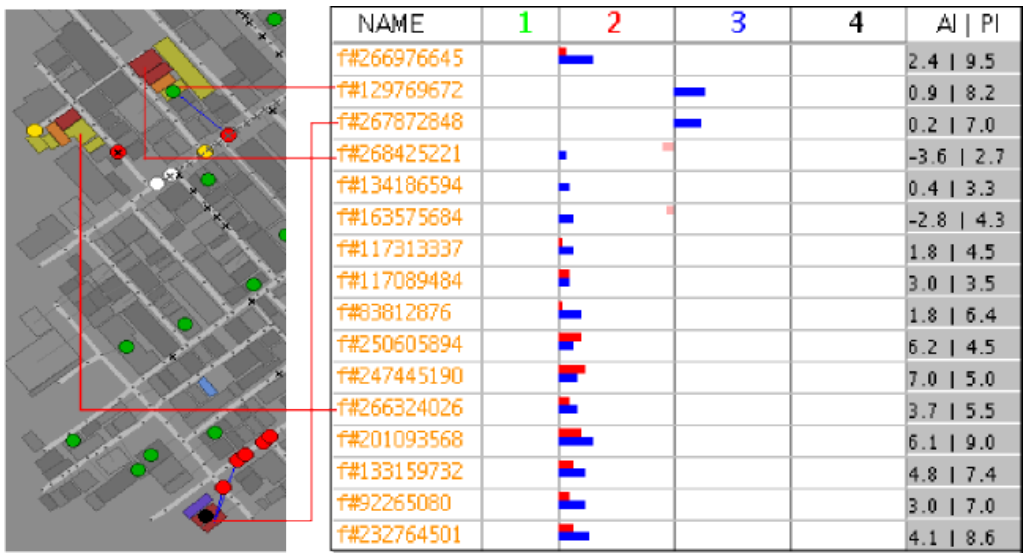

**Figure 9. Partial internal view of the representation MAS**

**CONCLUSION**

This paper presented a modeling of an agent-based DSS dedicated to the emergency management. We think the system is rather generic and that is possible to apply it on other types of applications. We choose essentially games as applications since we can have a complete knowledge about their environments and the rules are fixed in advance. At present, we work on two games addressed to the risk and the crisis management which are the game of Risk and the RCRSS. We proposed here a generic approach for modelling a crisis environment. The obtained model allows us to format extracted information into FSFs that will be managed by the first part of the DSS which is composed by factual agents and which intends to reflect the evolution of the current situation. The model was applied on RCRSS for which we defined a fine ontology in order to specify FSFs.

Some graphic tools we use for helping the decider (but also debugging in fact) are described in this paper. These tools help us understand the parameters of the factual agents which are the most accurate to characterise information and what are the essential data to transfer to the second layer of the global system.

In order to have an entire understanding of the overall behaviour of the system, we need to connect the representation layer with the two upper layers. Therefore, we are working currently at the same time on the characterisation and prediction layers. We started testing some cases and studying the evolution of the factual agents, what will give us an insight on the way we compare between the emergent clusters and the stored cases.